\documentclass{article}
\usepackage{spconf,amsmath,graphicx}

\usepackage{multirow}
\usepackage{hyperref}

\usepackage{amssymb}
\usepackage[dvipsnames, svgnames, x11names, table, xcdraw]{xcolor}

\usepackage{enumitem}
\setlist{nosep, leftmargin=14pt}
\usepackage{booktabs}
\usepackage{mwe} 

\title{UCAD: Uncertainty-guided Contour-aware Displacement for semi-supervised medical image segmentation}
\name{Chengbo Ding$^{1,2}$, Fenghe Tang$^{1,2}$, and Shaohua Kevin Zhou$^{1,2,3,4\star}$}

\address{   
$^{1}$ School of Biomedical Engineering, Division of Life Sciences and Medicine,\\
University of Science and Technology of China (USTC), Hefei, Anhui 230026, China\\
$^{2}$ Center for Medical Imaging, Robotics, Analytic Computing \& Learning (MIRACLE),\\
Suzhou Institute for Advanced Research, USTC, Suzhou, Jiangsu 215123, China\\
$^{3}$ Jiangsu Provincial Key Laboratory of Multimodal Digital Twin Technology,\\
Suzhou Jiangsu, 215123, China\\
$^{4}$ State Key Laboratory of Precision and Intelligent Chemistry, USTC, Hefei Anhui 230026, China
}
\begin{document}
%
\maketitle
\begin{abstract}
Existing displacement strategies in semi-supervised segmentation only operate on rectangular regions, ignoring anatomical structures and resulting in boundary distortions and semantic inconsistency. To address these issues, we propose \textbf{UCAD}, an \textit{\textbf{U}ncertainty-Guided \textbf{C}ontour-\textbf{A}ware \textbf{D}isplacement} framework for semi-supervised medical image segmentation that preserves contour-aware semantics while enhancing consistency learning. Our UCAD leverages superpixels to generate anatomically coherent regions aligned with anatomy boundaries, and an uncertainty-guided selection mechanism to selectively displace challenging regions for better consistency learning. We further propose a dynamic uncertainty-weighted consistency loss, which adaptively stabilizes training and effectively regularizes the model on unlabeled regions. Extensive experiments demonstrate that UCAD consistently outperforms state-of-the-art semi-supervised segmentation methods, achieving superior segmentation accuracy under limited annotation. The code is available at:
\href{https://github.com/dcb937/UCAD}{\textcolor{magenta}{\textit{https://github.com/dcb937/UCAD}}}.
\end{abstract}

\begin{figure}[t]
\includegraphics[width=8.5cm]{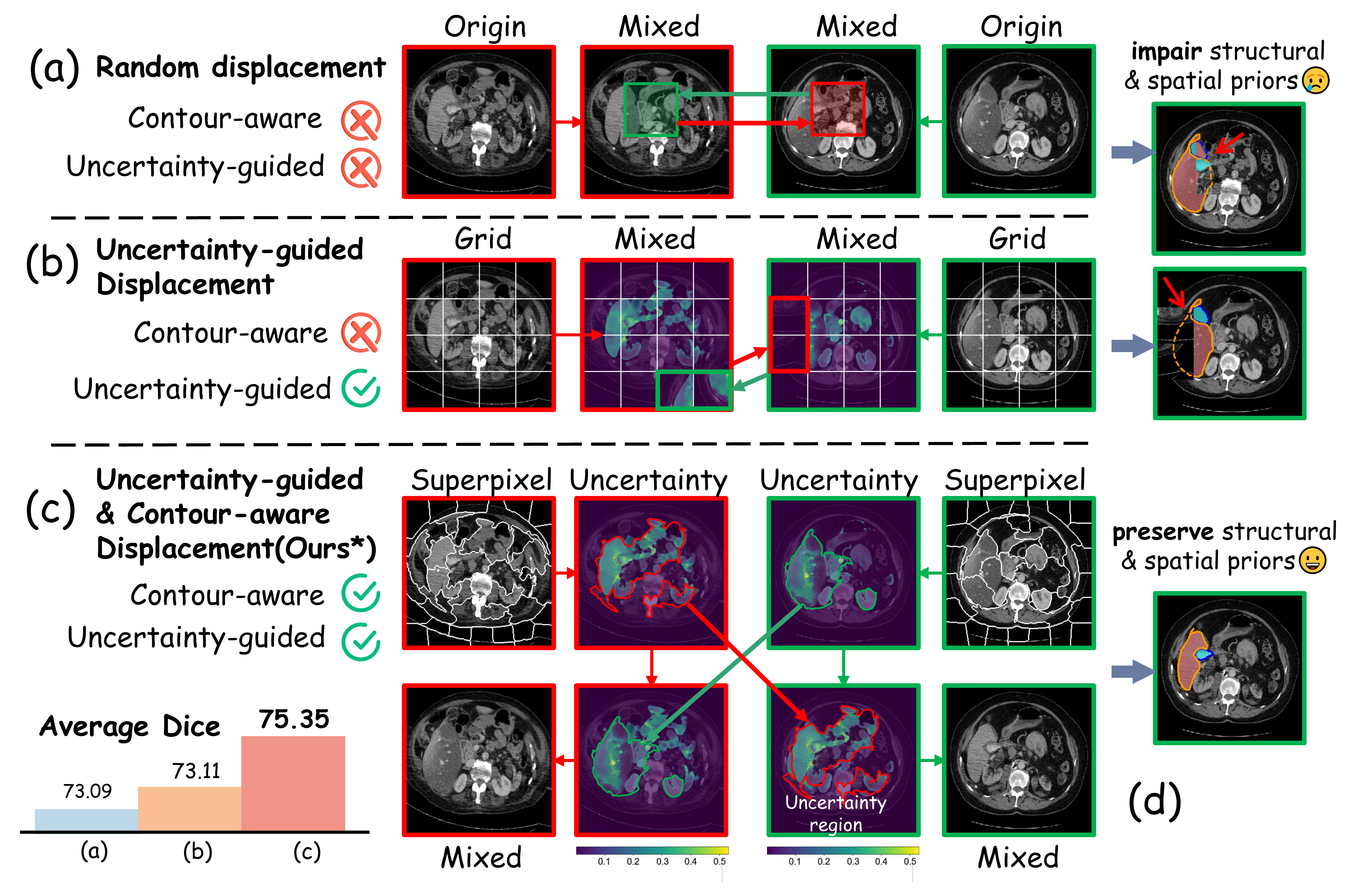}
\vspace{-4mm}
\caption{Illustration of different displacement strategies: 
(a) \textbf{Random displacement}: randomly exchanges rectangular regions between labeled and unlabeled images, ignoring anatomical boundaries; 
(b) \textbf{Uncertainty-guided displacement}: divides images into grid patches and performs uncertainty-guided region exchange, but still lacks structural alignment; 
(c) \textbf{Uncertainty-guided contour-aware displacement (ours)}: generates superpixel-based, anatomically coherent regions and selectively displaces uncertain superpixel regions; (d) Comparing displacement results, we observe that conventional methods often impair structural and spatial priors, while our method maintains this information.
}
\vspace{-5mm}
\label{fig:teasor}
\end{figure}

\section{Introduction}
\label{sec:intro}

Medical image segmentation is crucial yet heavily constrained by limited annotations~\cite{wang2023mcf,ubench,mobile,llm4seg,mambamim,hyspark}. Semi-supervised learning (SSL) has thus become a key approach to reduce labeling costs and enhance generalization~\cite{li2020shape}.
Most existing SSL methods can be categorized into three types: Input perturbations~\cite{tarvainen2017mean,yu2019uncertainty,bai2023bidirectional,luo2021semi,luo2021efficient}, model perturbations~\cite{wu2021semi,wang2023mcf,wu2024cross}, and feature perturbations~\cite{yang2025unimatch,mgcc}. 
Among them, input perturbations aim to enhance the model’s robustness by  altering the input images while preserving their semantic content~\cite{tarvainen2017mean}. 
In recent years, region-level mixing methods~\cite{bai2023bidirectional,wu2024cross,chi2024adaptive} have become representative approaches within this category.
Such strategies not only augment the data but also help alleviate the distribution mismatch between labeled and unlabeled samples~\cite{bai2023bidirectional}. 
From a broader view, these methods can be divided into \textbf{(i) Random displacement}, which performs random bidirectional copy–paste between labeled and unlabeled samples (e.g., BCP~\cite{bai2023bidirectional}, in Fig.~\ref{fig:teasor}(a)) and \textbf{(ii) Uncertainty-guided displacement}, partitions images into non-overlapping grids and conducts uncertainty-guided region exchange to encourage semantic complementarity (e.g., ABD~\cite{chi2024adaptive}, in Fig.~\ref{fig:teasor}(b)).

Despite their effectiveness, both random displacement and uncertainty-guided displacement share an inherent limitation: \textit{they operate exclusively on rectangular regions and neglect the importance of preserving shape information essential for accurate anatomy understanding}~\cite{raju2022deep} (Fig.~\ref{fig:teasor}(a, b, and d)). Anatomical structures in medical images typically exhibit irregular and curved boundaries. As shown in Fig.~\ref{fig:teasor}(d), when such rigid regions are copied or displaced, they may intersect anatomy contours unnaturally, leading to boundary distortions and semantic ambiguity. Moreover, uncertainty-guided rectangular regions displacement inadvertently relocates anatomically inconsistent regions, thereby disrupting semantic coherence and hindering the model’s ability to capture accurate structural and spatial representations. Such neglect restricts the model’s ability to capture shape-aware and semantically continuous representations, essential for accurate anatomical understanding. This motivates a key question: \textit{Can we retain these shape-aware semantics while still benefiting from region displacement strategies ?}

To address this challenge, we propose a novel \textit{\textbf{uncertainty-guided contour-aware displacement}} strategy, as shown in Fig.~\ref{fig:teasor}(c). Aiming to enhance the model’s contour-aware perception of anatomical structures, our method utilizes unsupervised superpixels to form spatially coherent, boundary-preserving regions that align with anatomical boundaries. By displacing these semantically meaningful regions between images, our approach generates mixed samples that preserve both the anatomical shape and contextual information of anatomy, ensuring that structural cues are maintained while enhancing data diversity. In addition, to identify which regions require displacement, we incorporate uncertainty estimation to selectively perturb ambiguous or low-confidence areas, typically near anatomical boundaries. By focusing perturbations on these high-uncertainty regions, the model is encouraged to refine vague contours and learn sharper boundary representations, leading to improved structural and spatial robustness in consistency learning. Moreover, we design a dynamic uncertainty-weighted consistency loss which adaptively stabilizes training and enhances meaningful regularization specifically for unlabeled regions. Extensive experiments demonstrate that our approach achieves state-of-the-art performance compared with recent methods.

\section{Method}
\label{sec:method}

\begin{figure}[t]

\begin{minipage}[b]{1.0\linewidth}
  \centerline{\includegraphics[width=8.5cm]{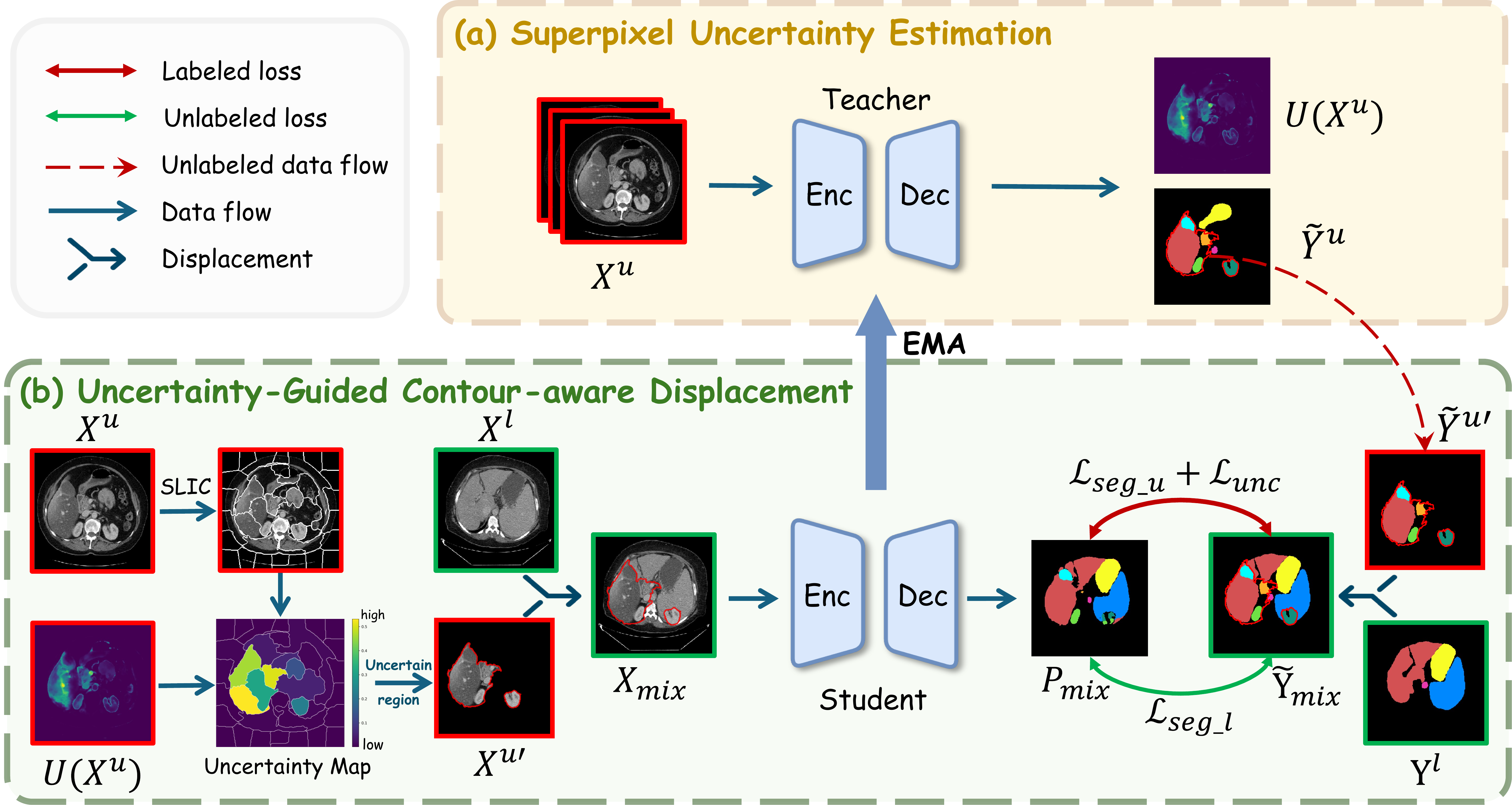}}
\end{minipage}

\caption{Overview of our framework.
(a) The teacher network takes the unlabeled image $X^u$ as input to generate a pseudo label $\tilde{Y}^u$ and a pixel-wise uncertainty score $U(X^u)$. 
(b) The unlabeled image $X^u$ is partitioned into superpixel region using the SLIC~\cite{achanta2012slic} algorithm, and the uncertainty $U(X^u)$ guides the selection of $N$ uncertain regions to form the mask $\mathcal{M}$. 
The masked regions are then displaced  with the labeled image $X^l$, resulting in a mixed image $X_{mix}$ and its corresponding mixed label $\tilde{Y}_{mix}$.
The student network is trained on $X_{mix}$ to predict $P_{mix}$, which is supervised by $\tilde{Y}_{mix}$ and an uncertainty-weighted consistency loss $\mathcal{L}_{unc}$.
For simplicity, the reverse displacement process is omitted.
}
\label{fig:framework}
\end{figure}

The overall framework of our proposed method is illustrated in Fig.~\ref{fig:framework}. 
Our approach is designed based on Mean Teacher~\cite{tarvainen2017mean}. 
We propose a contour-aware partition strategy for region-level mixing guided by superpixels, along with an uncertainty-driven displacement mechanism that prioritizes the replacement of difficult-to-learn regions to enhance supervision on challenging boundaries. Additionally, we employ a hybrid loss function that integrates supervised, consistency, and uncertainty-regularized terms.

\subsection{Uncertainty-Guided Contour-Aware Displacement}
\label{ssec:subhead}
\textbf{Contour-Aware Partition.}  
In order to improve the ability to capture anatomical contours under the semi-supervised setting, we introduce a contour-aware partition strategy guided by superpixels. The core idea is to replace selected superpixel regions from one image with the corresponding regions from another image, thereby generating mixed samples that preserve local structural consistency. Specifically, we adopt the Simple Linear Iterative Clustering (SLIC)~\cite{achanta2012slic} algorithm for creating superpixel regions:  
Let $\mathbf{X} \in \mathbb{R}^{H \times W}$ denote an input images with its corresponding labels $\mathbf{Y} \in \{0,1,\dots,C-1\}^{H \times W}$, where $C$ is the number of classes. Given a superpixel segmentation function $\mathcal{S}(\cdot)$, image $\mathbf{X}$ can be decomposed into a set of $K$ superpixel regions:
\begin{equation}
\mathcal{S}(\mathbf{X}) = \{ s_1, s_2, \dots, s_K \}, \quad \bigcup_{k=1}^K s_k = \Omega, \quad s_i \cap s_j = \emptyset \ (i \neq j),
\end{equation}
where $\Omega$ denotes the entire image domain and $s_k$ is a connected region of homogeneous appearance.

\noindent\textbf{Uncertainty-Guided Displacement.}  
Instead of sampling superpixel regions uniformly at random, we exploit the model's predictive uncertainty to determine which regions should be replaced during mixing. Specifically, given the probability map produced by the teacher network, we compute the pixel-wise Shannon entropy $H(\mathbf{X})$~\cite{shannon1948mathematical}.
For each superpixel $s$, the uncertainty score is obtained as the mean entropy within the region, and then transformed into a categorical distribution using a temperature-scaled softmax:
\begin{align}
U(s) &= \frac{1}{|s|} \sum_{x \in s} H(\mathbf{X}), \\
\mathcal{P}(s) &= \frac{\exp\left(U(s)/T\right)}{\sum_{s'} \exp\left(U(s')/T\right)} ,
\end{align}
where $T$ is the temperature parameter controlling the sharpness of the distribution, and $s' \in \mathcal{S}$ denotes all superpixels in the current image. 

Let $\mathcal{S}_{N} \sim \text{Categorical}(\mathcal{P}(s))$ denotes the set of the $N$ uncertain superpixels regions which is sampled according to the probability distribution $\mathcal{P}(s)$. The corresponding mask is generated by setting the selected regions to one while leaving other regions as zero:

\begin{equation}
\mathbf{M}(\mathbf{X}) = 
\begin{cases} 
1, & x \in \mathcal{S}_N, \\ 
0, & \text{otherwise}.
\end{cases}
\end{equation}

Then we define two input images $X_a$ and $X_b$ sampled from the labeled and unlabeled sets (in either order). 
Let $\mathcal{M} = \mathbf{M}(\mathbf{X}_a)$ be the selected mask. The mixed image $\tilde{\mathbf{X}}$ and its label $\tilde{\mathbf{Y}}$ can be generated as follows:
\begin{equation}
\tilde{\mathbf{X}} = \mathcal{M} \odot \mathbf{X}_a + (1 - \mathcal{M}) \odot \mathbf{X}_b,
\end{equation}
\begin{equation}
\tilde{\mathbf{Y}} = \mathcal{M} \odot \mathbf{Y}_a + (1 - \mathcal{M}) \odot \mathbf{Y}_b,
\end{equation}
where $\odot$ denotes the element-wise multiplication.
Our method leverages the structural prior provided by superpixels, ensuring that the mixed regions better align with object boundaries. 
Building upon this, the uncertainty-guided selection mechanism ensures that highly uncertain regions, which are more difficult to learn, are preferentially replaced during displacement, thereby enhancing structural supervision and enabling more accurate contour representation.

\subsection{The Overall Loss Function}
\label{ssec:loss}
\textbf{Region-Level Hybrid Loss.}  
To optimize the network, we adopt a hybrid loss function design that follows the design of BCP~\cite{bai2023bidirectional}. The overall objective integrates supervised loss on labeled regions, consistency loss on unlabeled regions, and an additional uncertainty-aware regularization.
Given a segmentation prediction $\tilde{\mathbf{Y}}$ for the mixed image $\tilde{\mathbf{X}}$, a labeled target $\mathbf{Y}_l$, a pseudo label $\mathbf{Y}_p$ for the unlabeled image, and a binary mask $\mathcal{M}$ indicating whether the region comes from the labeled image, we define a combinational loss as follows:
\begin{equation}
\begin{aligned}
\mathcal{L}_{seg} ={}& w_l \cdot \mathcal{L}_{DiceCE}(\tilde{\mathbf{Y}}, \mathbf{Y}_l; \mathcal{M}) \\
& + w_u \cdot \mathcal{L}_{DiceCE}(\tilde{\mathbf{Y}}, \mathbf{Y}_p; 1-\mathcal{M})
\end{aligned}
\end{equation}
where $\mathcal{L}_{DiceCE}$ is the linear combination of Dice loss and  Cross-entropy loss, and $w_l, w_u$ are weights for labeled and unlabeled regions, respectively.  

\noindent\textbf{Dynamic Uncertainty-Weighted Loss.}  
Inspired by the uncertainty-regularized consistency mechanism introduced in DyCON~\cite{assefa2025dycon}, we encourage the student prediction 
$\mathbf{p}_s$ to be consistent with the teacher prediction $\mathbf{p}_t$ to further regularize the model on unlabeled regions, while adaptively 
weighting the discrepancy according to predictive uncertainty. Specifically, the predictive entropy of both models, defined as the Shannon entropy over class probabilities, is used to adaptively weight the consistency discrepancy. The uncertainty-guided consistency loss is formulated as:
\begin{equation}  \label{unc}
\begin{aligned}
\mathcal{L}_{unc} = \frac{1}{\mathcal{N}_u} 
\sum_{x \in \Omega_u} \frac{\|\mathbf{p}_s(x) - \mathbf{p}_t(x)\|^2}{\exp(\beta H_s(x)) + \exp(\beta H_t(x))}  \\
+ \frac{\beta}{\mathcal{N}_u} \sum_{x \in \Omega_u} \big( H_s(x) + H_t(x) \big),
\end{aligned}
\end{equation}
where $\Omega_u$ denotes the unlabeled region, and $\mathcal{N}_u = |\Omega_u|$ is the number of voxels within it. 
$H_s(x)$ and $H_t(x)$ represent the predictive entropy of the student and teacher models.
The coefficient $\beta$ serves as an adaptive weighting factor that is gradually annealed during training.
Finally, the total loss is defined as the weighted sum of all components:
\begin{equation} \label{total_loss}
\mathcal{L}_{total} = \mathcal{L}_{seg} + \lambda \cdot \mathcal{L}_{unc},
\end{equation}
where $\lambda$ is a hyper-parameter balancing uncertainty regularization against supervised segmentation losses. This formulation ensures that the model benefits from both labeled supervision and unlabeled consistency.

\section{EXPERIMENTS AND RESULTS}
\label{sec:pagestyle}

\subsection{Experimental Setup}
\label{ssec:ExperimentalSetup}
\textbf{Dataset and Setting.}  
We conduct experiments on two 
publicly datasets: ACDC~\cite{bernard2018deep} and Synapse~\cite{landman2015miccai}. 
\textbf{ACDC:}  The Automated Cardiac Diagnosis Challenge (ACDC) dataset consists of 
cine-MRI scans acquired from 100 patients with different cardiac conditions. Following~\cite{bai2023bidirectional}, we use the same dataset split with 70 training, 10 validation, and 20 testing samples.
\textbf{Synapse:}  
The Synapse multi-organ CT dataset consists of 30 cases with 3,779 axial abdominal CT slices, each annotated for eight abdominal organs.
We split 18, 6, and 6 cases for training, validation, and testing, respectively.

\noindent\textbf{Implementation Details and Evaluation Metrics.}  
Following previous works~\cite{bai2023bidirectional,chi2024adaptive}, the input image size and patch size are set as 256×256 and the batch size is 24 during the training period. 
We use U-Net~\cite{ronneberger2015u} as the backbone networks and the SGD optimizer with weight decay of 1e-4, momentum of 0.9, learning rate of 0.01, followed same setting as~\cite{bai2023bidirectional,chi2024adaptive}. We conduct all experiments on an NVIDIA GeForce 4090 GPU with fixed random seeds. And we set $\lambda$ in Eq.~\ref{total_loss} as 0.2.
For evaluation, we utilize two widely used metrics: Dice Similarity Coefficient (DSC) and Average Surface Distance (ASD).

\begin{table}[t]
\centering
\begin{tabular}{lcccc}
\toprule
\multirow{2}{*}{Method} & \multicolumn{2}{c}{5\% label} & \multicolumn{2}{c}{10\% label} \\
\cmidrule(lr){2-3} \cmidrule(lr){4-5}
 & DSC(\%)$\uparrow$& ASD$\downarrow$& DSC(\%)$\uparrow$& ASD$\downarrow$ \\
\midrule
UA-MT~\cite{yu2019uncertainty}   & 46.04& 7.75& 81.65& 2.12 \\
SASSNet~\cite{li2020shape} & 57.77& 6.06& 84.50& 2.59 \\
DTC~\cite{luo2021semi} & 56.90& 7.39& 84.29& 2.36 \\
URPC~\cite{luo2021efficient} & 55.87& 3.74& 83.10& 2.28 \\
MC-Net~\cite{wu2021semi} & 62.85& 2.33& 86.44& 1.82 \\
SS-Net~\cite{wu2022exploring} & 65.83& 2.28& 86.78& 1.90 \\
MCF~\cite{wang2023mcf} & 82.37& 1.59& 87.67& 2.08 \\
BCP~\cite{bai2023bidirectional}& 87.59& 0.67& 88.84&1.17\\
\rowcolor{gray!15} UCAD (ours) & \textbf{88.63} & \textbf{0.49}&\textbf{89.93}&\textbf{0.57}\\
\bottomrule
\end{tabular}
\caption{Comparison results with SoTA semi-supervised segmentation methods on ACDC dataset.
Results of other methods are copied from BCP~\cite{bai2023bidirectional}, while the results of ours are reproduced under the same setting.}
\label{tab:acdc_results}
\vspace{-4mm}
\end{table}

\begin{table}[t]
\centering
\begin{tabular}{lcccc}
\toprule
\multirow{2}{*}{Method} & \multicolumn{2}{c}{5\% label} & \multicolumn{2}{c}{10\% label} \\
\cmidrule(lr){2-3} \cmidrule(lr){4-5}
 & DSC(\%)$\uparrow$& ASD$\downarrow$ & DSC(\%)$\uparrow$& ASD$\downarrow$ \\
\midrule

UA-MT~\cite{yu2019uncertainty}   & 31.96& 61.16& 40.63& 37.77\\
URPC~\cite{luo2021efficient} & 35.99& 38.98& 46.28& 27.35\\
MC-Net~\cite{wu2021semi} & 36.70& 52.13& 43.89& 35.78\\
CML~\cite{wu2024cross}& 36.52& 54.51& 58.49& 41.30\\
BCP~\cite{bai2023bidirectional}& 50.50& 55.89& 64.49&31.95\\
ABD~\cite{chi2024adaptive} & 51.28& 39.45& 62.41& \textbf{24.22}\\
\rowcolor{gray!15} UCAD (ours) & \textbf{56.10}& \textbf{34.77}& \textbf{66.73}&29.62\\
\bottomrule
\end{tabular}
\caption{Comparison results with SoTA semi-supervised segmentation methods on  Synapse dataset.}
\label{tab:Synapse_results}
\vspace{-4mm}
\end{table}

\begin{figure}[t]

\begin{minipage}[b]{1.0\linewidth}
  \centerline{\includegraphics[width=8.5cm]{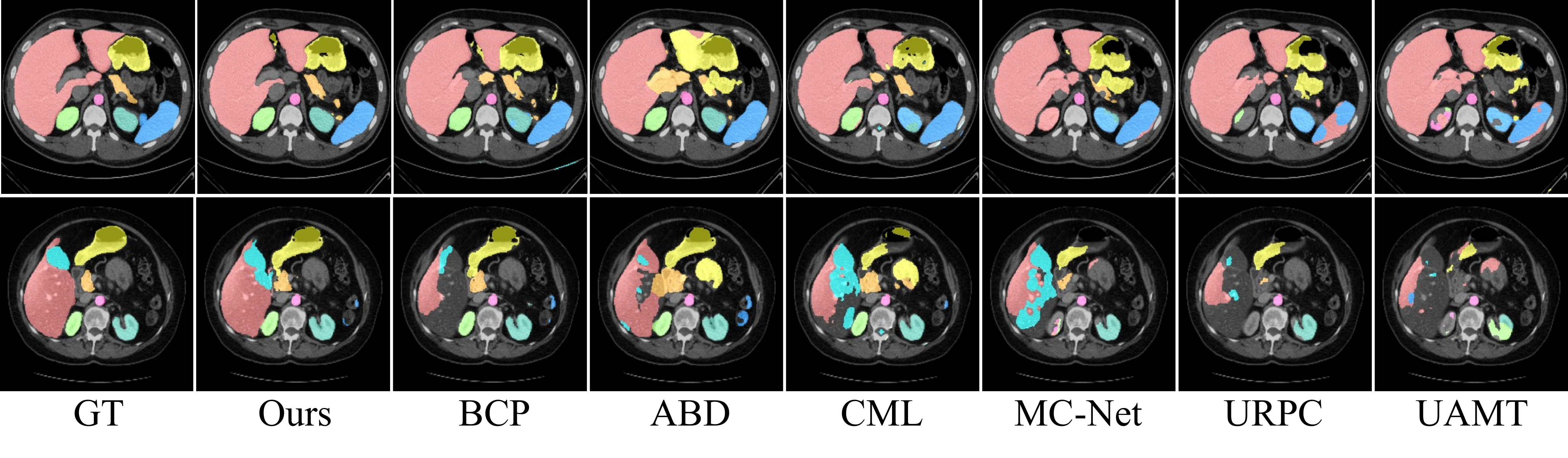}}
\end{minipage}
\vspace{-8mm}
\caption{Visual comparison on the 10\% Synapse dataset:
\textcolor[RGB]{0, 136, 255}{\rule{6pt}{6pt}}~spleen, 
\textcolor[RGB]{111, 225, 70}{\rule{6pt}{6pt}}~right kidney, 
\textcolor[RGB]{17, 144, 123}{\rule{6pt}{6pt}}~left kidney, 
\textcolor[RGB]{0, 255, 255}{\rule{6pt}{6pt}}~gallbladder, 
\textcolor[RGB]{252, 175, 42}{\rule{6pt}{6pt}}~pancreas, 
\textcolor[RGB]{212, 81, 83}{\rule{6pt}{6pt}}~liver, 
\textcolor[RGB]{248, 252, 39}{\rule{6pt}{6pt}}~stomach, 
and \textcolor[RGB]{234, 56, 175}{\rule{6pt}{6pt}}~aorta.}
\label{fig:visualization}
\end{figure}

\subsection{Comparison with State-of-the-art Methods}
\label{ssec:Comparison}

We compare the proposed method with several state-of-the-art (SoTA) semi-supervised medical image segmentation methods, including UA-MT~\cite{yu2019uncertainty}, SASSNet~\cite{li2020shape}, DTC~\cite{luo2021semi}, URPC~\cite{luo2021efficient}, MC-Net~\cite{wu2021semi}, SS-Net~\cite{wu2022exploring}, MCF~\cite{wang2023mcf}, BCP~\cite{bai2023bidirectional}, CML~\cite{wu2024cross}, and ABD~\cite{chi2024adaptive} under 5\% and 10\% labeled settings, as summarized in Tables~\ref{tab:acdc_results} and~\ref{tab:Synapse_results}.

As shown in Table~\ref{tab:acdc_results}, our UCAD consistently outperforms all competitors on the ACDC dataset under both label ratios. With 5\% labeled data, it achieves 88.63\% DSC and 0.49 ASD, outperforming previous SoTA methods and demonstrating effective use of unlabeled data. When the label ratio increases to 10\%, our method further improves to 89.93\% DSC and 0.57 ASD, showing stable and robust consistency learning.
Similarly, on the Synapse dataset (Table~\ref{tab:Synapse_results}), our approach achieves the best results under both 5\% and 10\% labeled settings, with DSC gains of 4.82\% and 2.24\% over ABD. The significant improvement on complex multi-organ CT scans highlights the advantage of our contour-aware mixing in preserving anatomical structures and refining boundaries.

\subsection{Ablation Study}
\label{ssec:Ablation}
We conduct ablation experiments on the Synapse dataset using 10\% labeled data to evaluate the contribution of each proposed component, as summarized in Table~\ref{tab:Ablation}. Starting from the baseline model that adopts standard rectangular regions for displacement, introducing the contour-aware displacement (CAD) improves anatomical alignment and enhances boundary accuracy. Adding uncertainty-guided superpixel selection (UGS) further boosts performance by directing the displacement operations toward regions with low prediction confidence. Finally, integrating the uncertainty-weighted consistency loss $\mathcal{L}_{unc}$ (Eq.~\ref{unc}) provides additional training stability by adaptively down-weighting unreliable pseudo-labels. The complete model achieves the best overall performance with a Dice score of 66.73\% and an ASD of 29.62, demonstrating that all components work synergistically to achieve robust and shape-consistent semi-supervised segmentation.

\begin{table}[t]
\centering
\begin{tabular}{cccc|cc}
\hline
Base & CAD & UGS & $\mathcal{L}_{unc}$ & DSC$\uparrow$ & ASD$\downarrow$ \\
\hline
$\checkmark$ &  &  &  & 64.49 & 31.95\\
$\checkmark$ & $\checkmark$ &  &  &  65.63&   32.38\\
$\checkmark$ & $\checkmark$ & $\checkmark$ &  &  66.38&   29.69\\
$\checkmark$ & $\checkmark$ & $\checkmark$ & $\checkmark$ & \textbf{66.73}&  \textbf{29.62}\\
\hline
\end{tabular}
\caption{Ablation study on Synapse dataset with 10\% labeled data. 
"Base" means using standard rectangular regions for displacement as the baseline. 
CAD means contour-aware displacement. 
UGS means adding uncertainty-guided selection in displacement. 
$\mathcal{L}_{unc}$ means uncertainty-weighted consistency loss defined in Eq.~\ref{unc}.}
\label{tab:Ablation}
\vspace{-4mm}
\end{table}

\section{CONCLUSION}
\label{sec:CONCLUSION}

In this paper, we proposed a novel uncertainty-guided contour-aware displacement framework UCAD. UCAD uses uncertainty to guide superpixels in generating anatomically aligned regions, incorporating shape awareness into input-level perturbations, and employs a dynamic uncertainty-weighted consistency loss to adaptively emphasize ambiguous regions and stabilize the learning process. Overall, UCAD achieves remarkable results across extensive experiments, it provides a promising solution for semi-supervised medical image segmentation, with the potential to significantly improve segmentation accuracy under limited annotations.

\section{Compliance with ethical standards}
\label{sec:ethics}
Informed consent was obtained from all individual participants involved in the study.

\section{Acknowledgments}
\label{sec:acknowledgments}

This work was supported by Natural Science Foundation of China under Grant 62271465 and Suzhou Basic Research Program under Grant SYG202338.

\bibliographystyle{IEEEbib}
\bibliography{strings,refs}

\end{document}